\documentclass[10pt, conference, compsocconf]{IEEEtran}
\IEEEoverridecommandlockouts
\usepackage{cite}
\usepackage{amsmath,amssymb,amsfonts}
\usepackage{algorithmic}
\usepackage{graphicx}
\usepackage{textcomp}
\usepackage{xcolor}
\usepackage{hyperref}
\usepackage{xurl}
\usepackage{subcaption}

\def\BibTeX{{\rm B\kern-.05em{\sc i\kern-.025em b}\kern-.08em
    T\kern-.1667em\lower.7ex\hbox{E}\kern-.125emX}}
\begin{document}

\title{Service-based Trajectory Planning in Multi-Drone Skyway Networks}

\author{\IEEEauthorblockN{Sarah Bradley\IEEEauthorrefmark{1},
Albertus Alvin Janitra\IEEEauthorrefmark{2}, Babar Shahzaad\IEEEauthorrefmark{3},
Balsam Alkouz\IEEEauthorrefmark{4},
Athman \\Bouguettaya\IEEEauthorrefmark{5}, and
Abdallah Lakhdari\IEEEauthorrefmark{6}}
\IEEEauthorblockA{School of Computer Science,
The University of Sydney\\
Australia\\
Email: \IEEEauthorrefmark{1}sbra0523@uni.sydney.edu.au,
\IEEEauthorrefmark{2}ajan8924@uni.sydney.edu.au,
\IEEEauthorrefmark{3}babar.shahzaad@sydney.edu.au,\\
\IEEEauthorrefmark{4}balsam.alkouz@sydney.edu.au,
\IEEEauthorrefmark{5}athman.bouguettaya@sydney.edu.au,
\IEEEauthorrefmark{6}abdallah.lakhdari@sydney.edu.au
}}






\maketitle

\begin{abstract}
We present a demonstration of service-based trajectory planning for a drone delivery system in a multi-drone skyway network. We conduct several experiments using Crazyflie drones to collect the drone's position data, wind speed and direction, and wind effects on voltage consumption rates. The experiments are run for a varying number of recharging stations, wind speed, and wind direction in a multi-drone skyway network. \\ Demo: \url{https://youtu.be/zEwqdtEmmiw}  
\end{abstract}

\begin{IEEEkeywords}
Drone Delivery, Flight Trajectory, Drone Service, Multi-Drone Skyway Network.
\end{IEEEkeywords}

\section{Introduction}

\begin{figure}[t]
\centering
\includegraphics[width=0.75\linewidth]{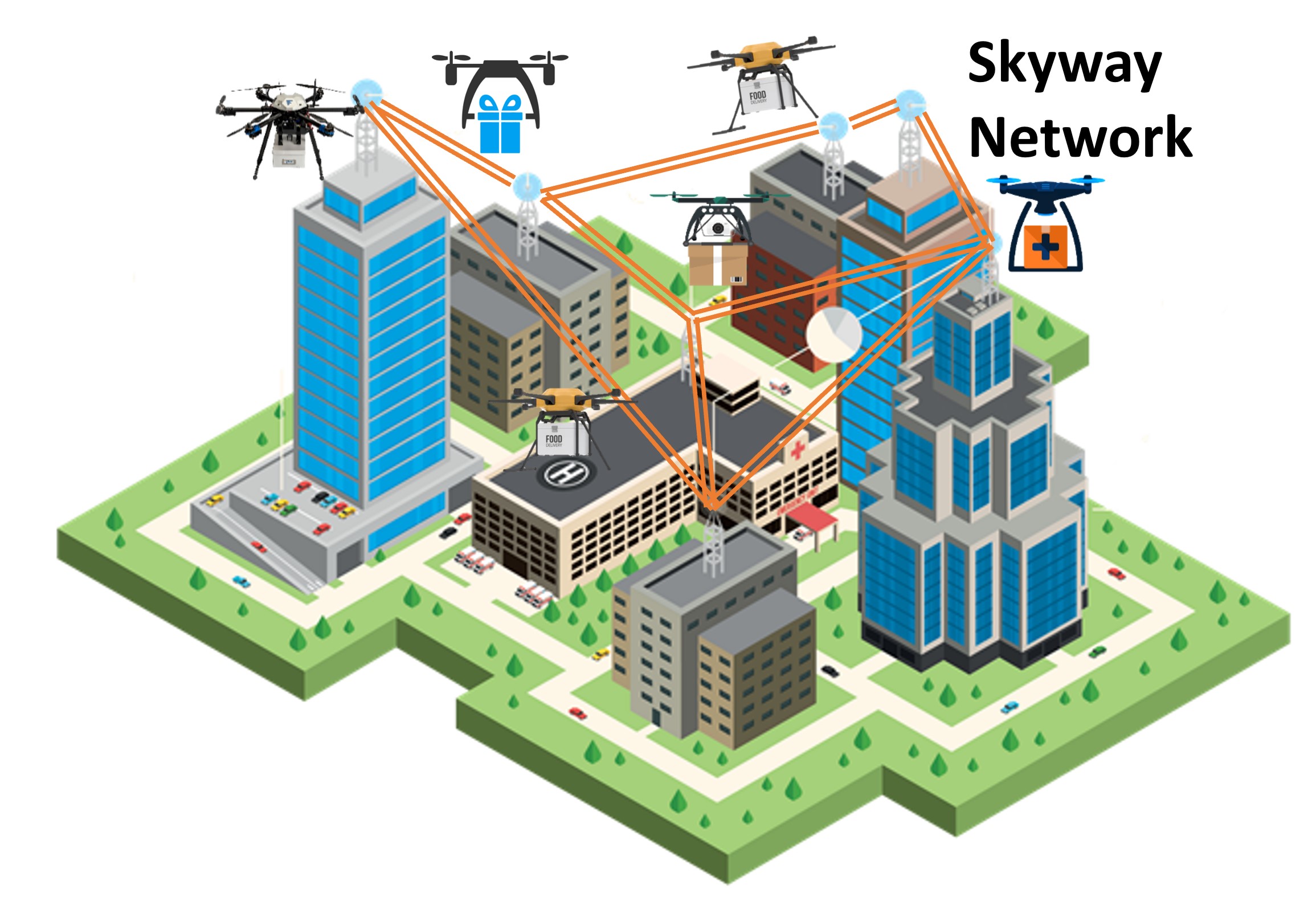}
\caption{Multi-Drone Skyway Network}
\label{skyway}
\vspace{-0.6 cm}
\end{figure}

Drones are a specific type of unmanned aerial vehicles that fly \textit{autonomously} with full network connectivity capabilities \cite{8119717}. This connectivity enables drones to operate \textit{safely} and provide a multitude of civilian applications on large scales \cite{9885785}. Examples of these applications include geographic mapping and surveying, disaster management, and delivery of packages \cite{shakhatreh2019unmanned}. The \textit{ubiquity} of drones in the sky has prompted an increasing interest of several logistics companies such as Amazon, Wing, and Flytrex to use drones for package delivery \cite{alkouz2022density}. Drones provide \textit{faster, safer, contactless}, and \textit{resilient} delivery solutions \cite{10016378}. However, drones have \textit{inherent constraints} on battery capacity, flight range, and payload capacity \cite{10.1145/3460418.3479289}. These constraints hinder the full potential utilization of drones in dynamic environments \cite{alkouz2021service}. The deployment of a \textit{multi-drone skyway network} addresses these constraints and enables a robust service delivery in a dynamic environment with shared airspace \cite{wang2017integrating,9590339,wang2009web}. A multi-drone skyway network is constructed by linking a set of skyway segments where nodes are the rooftops of high-rise buildings \cite{liu2022constraint}. We concurrently use these rooftops both as recharging stations and delivery targets (Fig. \ref{skyway}).

This demo presents a novel \textit{service-based trajectory planning} for the provisioning of drone delivery in a multi-drone skyway network. Our service-based trajectory planning considers the recharging constraints of drones during the delivery operation. In this respect, these drones may need to wait for recharging pad availability if they share the \textit{same station}. This wait for recharging pad availability directly impacts the overall delivery time \cite{shahzaad2019constraint,shahzaad2021top}. Wireless recharging pads at intermediate stations enable and support long-distance deliveries \cite{jermaine2021demo}. We collect a multi-drone dataset under a \textit{range of parameters} where the drones wait for the recharging pad availability to be recharged. The parameters recorded in this dataset include the impact of varying wind conditions on the drone's battery consumption. We believe this dataset constitutes a valuable resource for understanding the drone's \textit{battery consumption behavior} and planning the \textit{time-optimal deliveries}.





\begin{figure}[h]
  \centering
  \includegraphics[width=0.78\linewidth]{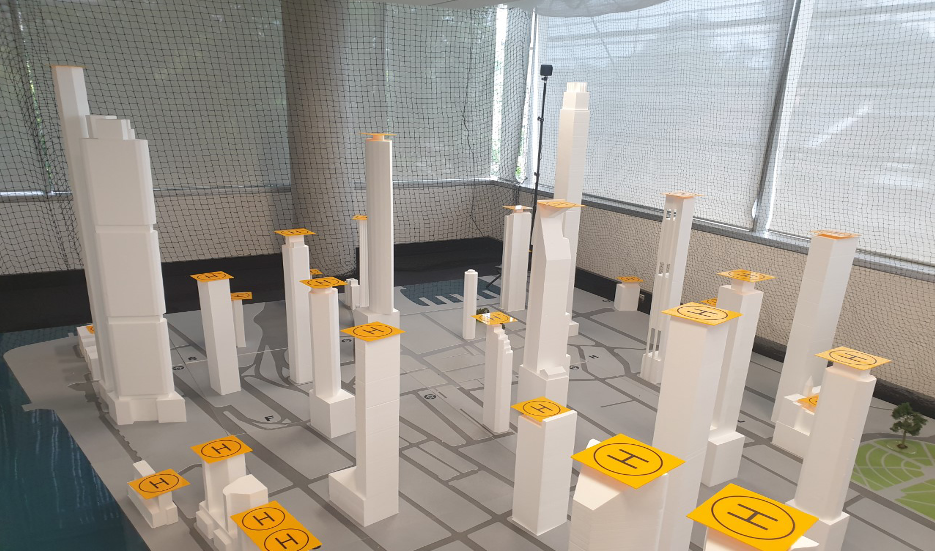}
  \caption{3D Model of City CBD}\label{3d}
\vspace{-0.4 cm}
\end{figure}

\begin{figure}[t]
\centering
  \includegraphics[width=0.8\linewidth, height= 5.5cm]{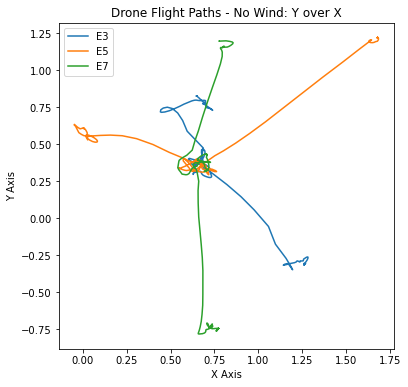}
  \caption{Single Station Flight Path}
\label{flightpath1}
\vspace{-0.4 cm}
\end{figure}

\section{Demo Setup}


We use three Crazyflie 2.1 nano-quadcopter drones by Bitcraze in a 3D model of the City CBD as an indoor testbed to mimic a multi-drone skyway network (Fig. \ref{3d}). Two HTC Vive base stations are used that employ infrared laser technology to calculate the coordinates of each drone. A fan with adjustable speed is used to simulate the wind conditions. Two NORDMÄRKE wireless chargers\footnote{\url{https://www.ikea.com/au/en/p/nordmaerke-wireless-charger-white-cork-90478064/}} are utilized for wireless recharging of drones at the intermediate stations.

\section{Data Collection}

We use Crazyflie Python API for data collection with the settings of single and two intermediate recharging stations. We performed 70 flights from which 35 flights with a \textit{single station} (Fig. \ref{flightpath1}) and 35 with \textit{two stations} (Fig. \ref{flightpath2}). Each drone records the following data points with a timestamp of 100 ms during its flight: (x, y, z) coordinates, roll, pitch, yaw, voltage, relative wind direction, travel distance, and waiting time at a recharging station.
The data collected from the experiments are organized in CSV files by the number of recharging nodes, path, wind speeds, wind direction, and drone. 
The drones are labeled as E3, E5, and E7 based on the last two characters in their unique radio addresses. Additionally, we track the change in voltage of all three drones while recharging on a wireless recharging pad from low voltage (3.2 volts) to full voltage (greater than 4.15 volts). We consider a number of operational parameters (wind speeds, wind directions, skyway segments, initial voltage) to fly each drone.



\subsection{Varying Number of Recharging Stations}
\label{path composition}
This section describes three drones operating simultaneously in a skyway network with a single and two stations.
\subsection{Single Recharging Station}
In the case of a single recharging station, each drone starts from a unique source node, flies to an intermediate recharging station, and finally moves to a unique destination node (Fig. \ref{flightpath1}). Fig. \ref{flightpath1} shows the flight paths with 1 intermediate recharging station, with the cluster at (0.6, 0.3) as the intermediate recharging node. The drones fly to the intermediate recharging node in order of closest to farthest from the intermediate node. The order of the drones is decided autonomously during the experiment, using the drone's coordinates to calculate their distance from the intermediate node. Upon arriving at the intermediate node, a drone recharges until reaching a threshold of 4.15 volts. After recharging, the next drone flies to the intermediate node, while the first drone flies to its destination node. We conduct this experiment with five path sets (a total of 5 paths for each drone, 15 paths in total). Each path is tested under different wind conditions of no wind, 6.1 km/h wind, and 7.6 km/h wind. Each test is conducted under three different wind directions based on the fan's location, including North, South, and East winds relative to the City's 3D model. 

\begin{figure}[t]
\centering
  \includegraphics[width=0.8\linewidth, height= 5.5cm]{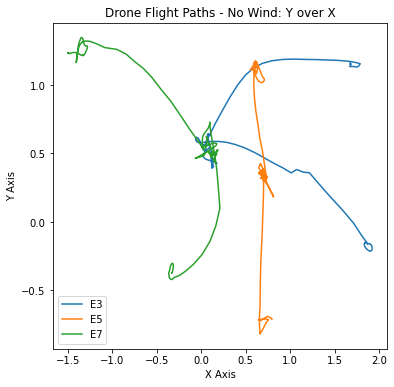}
\caption{Two Stations Flight Path}
\label{flightpath2}
\vspace{-0.4 cm}
\end{figure}



\begin{figure*}[t]
  \centering
  \begin{subfigure}[t]{0.5\linewidth}
    \includegraphics[width=\linewidth]{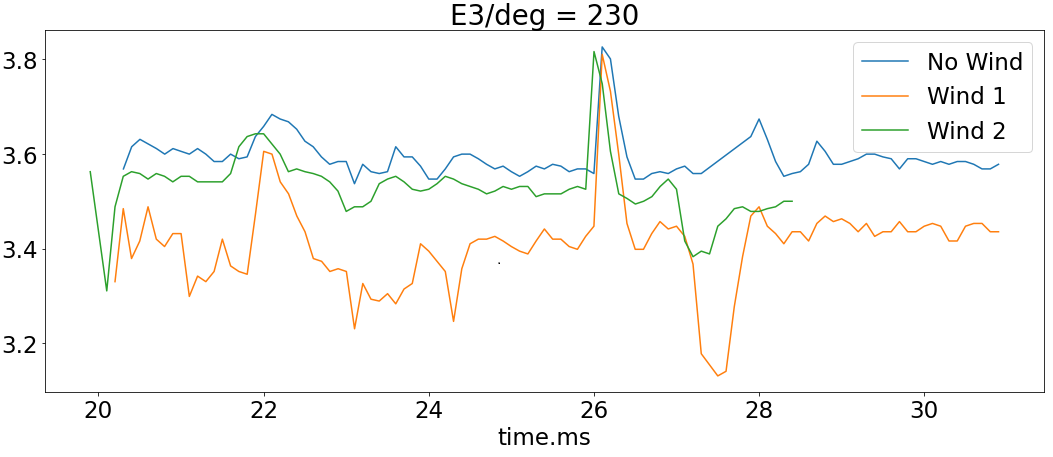}
  \vspace{-0.4cm}
   \caption{}\label{rectangle-diagram}
  \end{subfigure}\hfill
  \begin{subfigure}[t]{0.5\linewidth}
    \includegraphics[width=\linewidth]{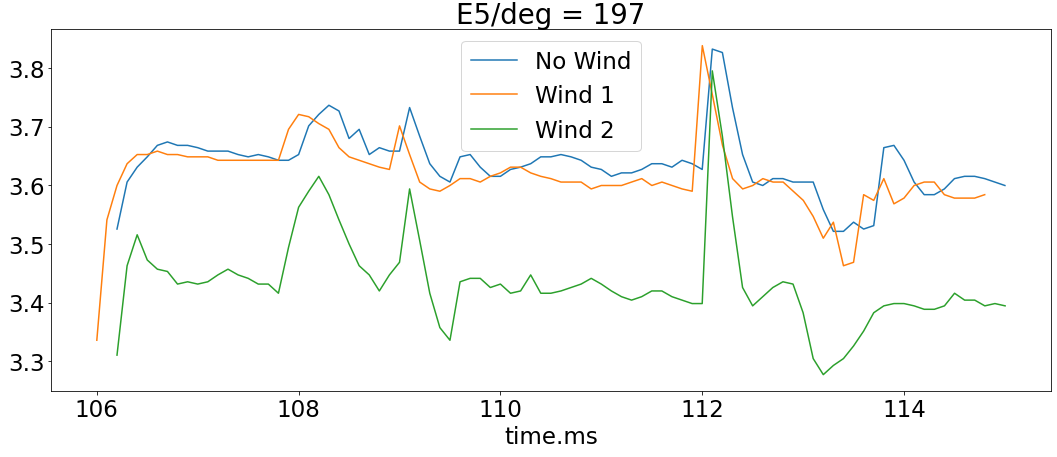}
  \vspace{-0.4cm}
   \caption{}\label{triangle-diagram}
  \end{subfigure}
  \begin{subfigure}[t]{0.5\linewidth}
    \includegraphics[width=\linewidth]{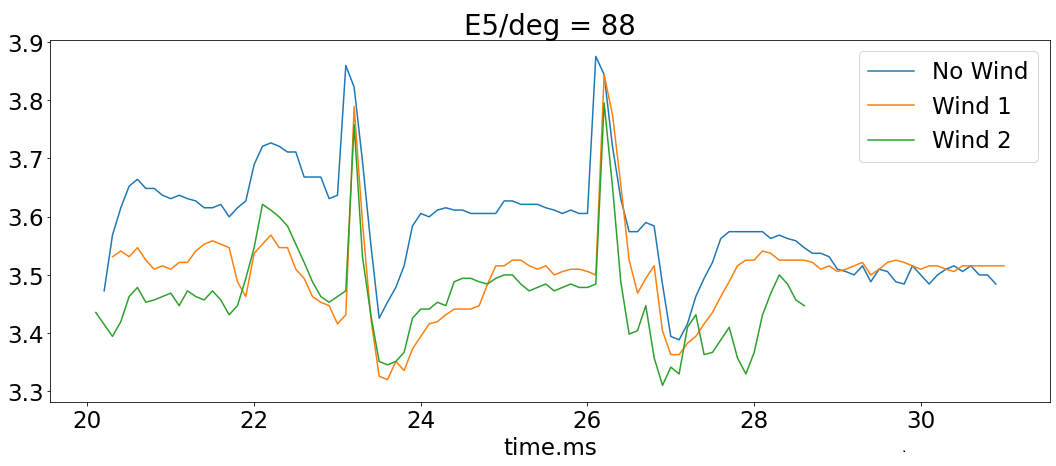}
  \vspace{-0.4cm}
    \caption{}\label{line-diagram}
  \end{subfigure}
  \caption{Wind Speed and Direction Impact on Voltage Consumption}
  \label{fig:voltage}
  \vspace{-0.4cm}
\end{figure*}

\subsection{Two Recharging Stations}

In the case of two recharging stations, a drone progresses to one of the two intermediate recharging nodes and travels to its final destination node after recharging (Fig. \ref{flightpath2}). Fig. \ref{flightpath2} shows the flight paths with 2 intermediate recharging nodes with the clusters at (0.0, 0.5) and (0.7, 0.4) representing the two intermediate recharging nodes. We record the effects of wind on drone battery consumption. As there are two intermediate recharging nodes (A and B), the drone closest to intermediate recharging node A flies to A, and the drone closest to intermediate recharging node B flies to B. The intermediate recharging node allocated to each drone, and the order in which the drones fly, are decided autonomously at run time. This decision is made using our nearest first algorithm, which calculates the distances of each drone from the intermediate nodes and sends real-time instructions to the drones for the next node. When the first drone reaches a voltage of 4.15, it progresses to its destination node. This progression triggers the third drone to fly to the next available recharging station. We conduct this experiment with five similar path sets similar to the single recharging station. Each path has unique start, intermediate, and destination nodes for each drone. We use similar wind conditions and wind directions to test each path. We observe certain deviations from a straight line path due to occasional delays in calibrating the drone's positioning system.




\section{Results}

Fig. \ref{fig:voltage} shows the voltage over time during the drone flight from one node to another. Each plot shows the same skyway segment traversed by a drone under three different wind conditions: no wind, 6.1 km/h wind, and 7.6 km/h wind. The overall contour of voltage for each segment is similar within each plot. However, the segments under \textit{windy conditions} show a more \textit{irregular contour}. This irregularity could be due to the wind effects that require a drone to consume extra power to stay on the correct path.
This extra power drain and the added turbulence to the drone with no wind tend to start at a higher voltage on average and maintain a higher voltage throughout the segment.
The variations are also due to the different relative wind directions \cite{shahzaad2022drone}. For example, a drone flying into the \textit{headwind} requires more power than a drone flying in the same direction as the wind (i.e., \textit{tailwind}) \cite{lee2022autonomous}. Therefore, we observe that wind 2 consumes more voltage compared to wind 1 in Fig. \ref{rectangle-diagram} and wind 1 consumes more voltage than wind 2 in Figs \ref{triangle-diagram} and \ref{line-diagram}. This increase in voltage consumption is due to the change in wind direction that affects the lifting force of the drone.

\section {Demonstration Requirements}
This demonstration will present service-based trajectory planning in multi-drone skyway networks. The Crazyflie drones will use the HTC Vive base stations to locate themselves in an indoor drone testbed. The drones will be recharged using wireless pads to serve long-distance service delivery. We will show a pre-recorded video of this demonstration during the conference, which is available at \url{https://youtu.be/zEwqdtEmmiw}. The indoor drone testbed is very large in size and features a safety net for drones, as shown in the video. Therefore, we plan to do a live stream for the demonstration with one of our authors from our drone lab. We will use our laptops for this live stream. However, it will require a standard booth setup at the conference venue. A screen projector is preferable to help in presenting a better live stream.

\section*{Acknowledgment}

This research was partly made possible by LE220100078 grant from the Australian Research Council. The statements made herein are solely the responsibility of the authors.

\bibliographystyle{IEEEtran}  
\bibliography{percom} 

\end{document}